\documentclass{bmcart}

\usepackage{amsthm,amsmath}
\RequirePackage{hyperref}
\usepackage[utf8]{inputenc} 

\usepackage{tabularx,ragged2e}
\newcolumntype{C}{>{\Centering\arraybackslash}X} 

\usepackage{graphicx}

\startlocaldefs
\endlocaldefs

\begin{document}

\begin{frontmatter}

\begin{fmbox}
\dochead{Research}


\title{A Conversational Agent System for Dietary Supplements Use}


\author[
  addressref={aff1},                   
  email={sing0640@umn.edu}   
]{\inits{E.S}\fnm{Esha} \snm{Singh}}
\author[
  addressref={aff2},
  email={bompe001@umn.edu}
]{\inits{A.B}\fnm{Anu} \snm{Bompelli}}
\author[
  addressref={aff1},
  email={wanxx199@umn.edu}
]{\inits{A.Y}\fnm{Ruyuan} \snm{Wan}}
\author[
  addressref={aff4}, 
  email={bianjiang@ufl.edu}
]{\inits{J.B}\fnm{Jiang} \snm{Bian}}
\author[
  addressref={aff2},
  email={pakh0002@umn.edu}
]{\inits{S.P}\fnm{Serguei} \snm{Pakhomov}}
\author[
  addressref={aff3},
  corref={aff3}, 
  email={zhan1386@umn.edu}
]{\inits{R.Z}\fnm{Rui} \snm{Zhang}}



\address[id=aff1]{
  \orgdiv{CSE Department, College of Science and Engineering},             
  \orgname{University of Minnesota},          
  \city{Minneapolis, MN},                              
  \cny{US}                                    
}
\address[id=aff2]{%
  \orgdiv{College of Pharmacy},
  \orgname{University of Minnesota},
  \city{Minneapolis, MN},
  \cny{US}
}

\address[id=aff3]{%
  \orgdiv{Institute for Health Informatics and College of Pharmacy},
   \orgname{University of Minnesota},
  \city{Minneapolis, MN},
  \cny{US}
}
\address[id=aff4]{%
  \orgdiv{Health Outcomes \& Biomedical Informatics, College of Medicine},
   \orgname{University of Florida},
  \city{Gainesville, FL},
  \cny{US}
}


\end{fmbox}


\begin{abstractbox}

\begin{abstract} 
\parttitle{Background} 
Dietary supplements (DS) have been widely used by consumers, but the information around the efficacy and safety of DS is disparate or incomplete, thus creating barriers for consumers to find information effectively. Conversational agent (CA) systems have been applied to healthcare domain, but there is no such a system to answer consumers regarding DS use, although widespread use of DS. In this study, we develop the first CA system for DS use.

\parttitle{Methods} 
Our CA system for DS use developed on the MindeMeld framework, consists of three components: question understanding, DS knowledge base, and answer generation. We collected and annotated 1509 questions to develop natural language understanding module (e.g., question type classifier, named entity recognizer) which was then integrated into MindMeld framework. CA then queries the DS knowledge base (i.e., iDISK) and generates answers using rule-based slot filling techniques. We evaluated algorithms of each component and the CA system as a whole.

\parttitle{Results} 
CNN is the best question classifier with F1 score of 0.81, and CRF is the best named entity recognizer with F1 score of 0.87. The system achieves an overall accuracy of 81\% and an average score of 1.82 with succ@3+ score as 76.2\% and succ@2+ as 66\% approximately. 

\parttitle{Conclusion} 
This study develops the first CA system for DS use using MindMeld framework and iDISK domain knowledge base.
\end{abstract}




\end{abstractbox}
%

\end{frontmatter}



\section{Introduction}
The utilization of DS (e.g., vitamins, minerals, botanical extracts, and protein powders) in the United States (US) has dramatically increased in recent years. The 2019 Council for Responsible Nutrition (CRN) survey shows that 77\% of US adults take DS and 87\% (5\% increase in 2 years) express overall confidence in the safety, quality and effectiveness of DS 
\cite{nih_factsheets}. The utilization of DS does not require prescription, and consumers usually find health information about DS use by searching the internet themselves. Many sites contain basic facts about DS, their therapeutic use, safety warnings, effectiveness, and information on DS-related research studies. However, a lot of this online DS information is of low-quality; and sources of this information are also heterogeneous in nature, ranging from health organizations, government agencies, universities, interest groups, and lay consumers. The distribution of DS information across various DS resources (e.g., Natural Medicines \cite{nm}, Memorial Sloan Kettering Cancer Center \cite{mskdcc_website}) was found fragmented and inconsistent. Further, the knowledge representations used in these DS resources vary from unstructured free text to more structured searchable data. While these databases or resources provide basic DS knowledge, they either contain incomplete information or lack a standardized knowledge representation (e.g., not using standardized terms for adverse events) that allows these resources to be integrated into a more comprehensive DS KB. Recently, we have developed an integrated and standard DS knowledge base (i.e., iDISK \cite{idisk}), which can facilitate efficient and meaningful dissemination of DS knowledge. In the study, our system is developed with the assistance of iDISK as the DS knowledge base in the backend. 
\par
There is extensive prior work on natural language understanding and answering consumer questions regarding various health related issues \cite{liu2011},\cite{robert2016},\cite{eagli_website} and a number of automated online and offline biomedical conversational systems exist \cite{loqda},\cite{florence},\cite{clue},\cite{getforksy},\cite{woebot},\cite{abbi}. Recent development and advances in voice recognition, natural language processing (NLP), and artificial intelligence (AI) have led to the increasing availability and use of conversational agents (CA) - any dialogue system that not only conducts NLP but also responds automatically using human language \cite{conversation_ai}. Thus, CA have increasingly become an integral part of our day-to-day lives. CA systems could be classified into 4 types: a) interaction mode, b) chatbot application, c) rule-based AI, d) domain specific or open domain \cite{survey}. Based on their goals, CAs can be categorized into two main types a) task-oriented and b) non-task-oriented \cite{ca_limitations}\cite{survey_ai}\cite{yan2017}. Recent work on CAs has focused on personalization of CA \cite{spillman} as well as CA applications in specific domains \cite{morbini2012}. There has also been some work on CAs in the biomedical domain \cite{kimani2016}\cite{dinsa2018}\cite{kocaballi2019} with one of most recent publication by Dina et al. \cite{dina}.
\par
To the best of our knowledge, there is no prior study on the development of a CA system for DS consumers. Thus, the objective of this study is to develop a DS-focused CA system to answer questions related to DS use. We also evaluated the feasibility of using the domain knowledge base (i.e., iDISK \cite{idisk}) and the MindMeld framework for developing the CA system. Thus, our paper describes a task-oriented and domain specific CA system.

\section{Methodology}
The system architecture and the details of specific processes in each step of its development are discussed in this section. We also discuss evaluation process for our CA system . In the following sections, we described 1) the CA system architecture 2) dataset and annotation, 3) knowledge base, 4) methods for understanding users' questions, 5)answer generation, and 6) Evaluation of the CA system.

\begin{figure}[ht!]
\includegraphics[scale=0.35]{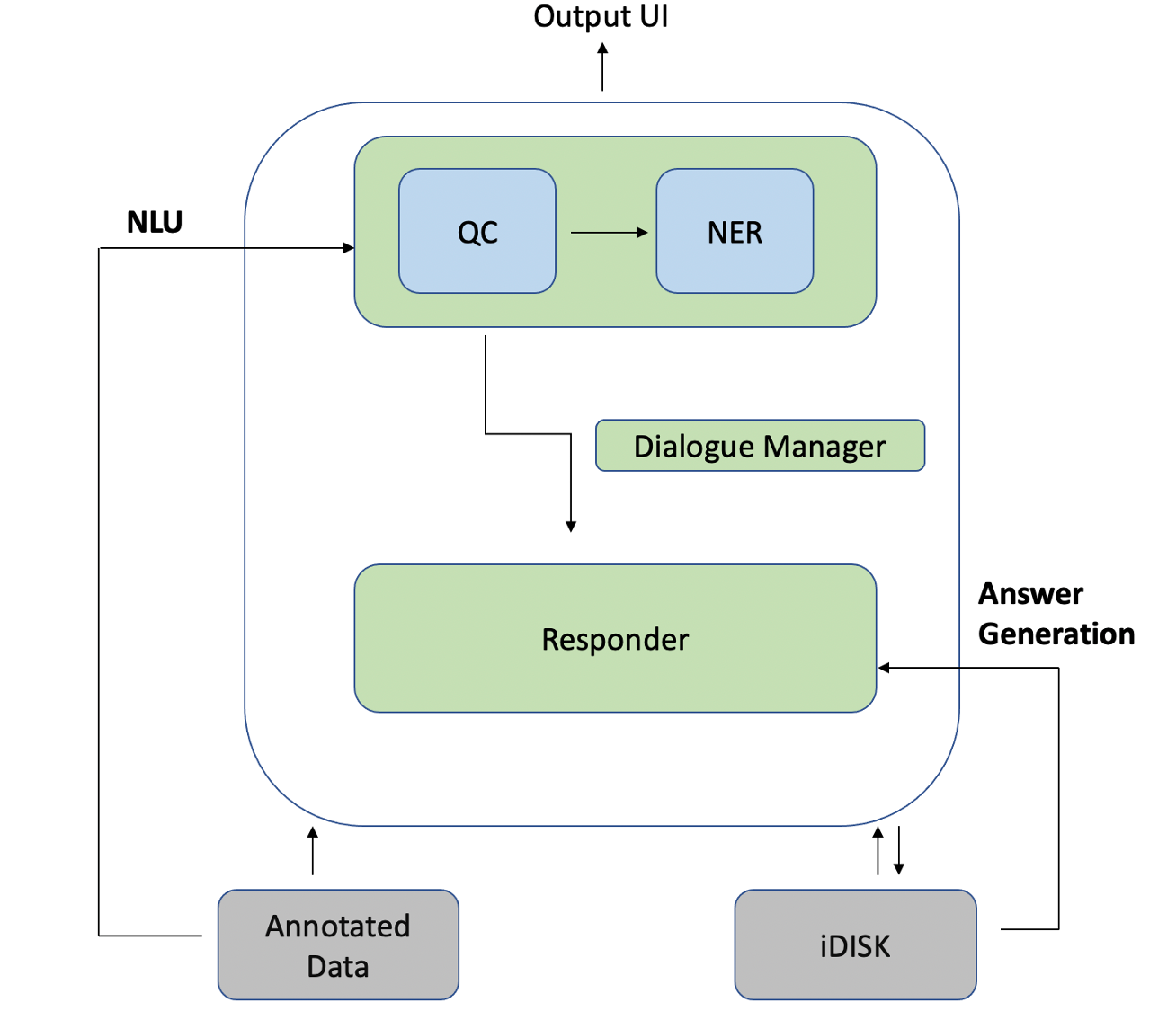} 
  \caption{CA system Architecture}
  \label{figure1}
\end{figure}

\subsection{CA System Architecture}

We chose the MindMeld as our CA system platform. MindMeld is very well designed for task-driven CA applications and has a set of robust utilities for deploying CA applications as a web service.  Compared to other CA systems (e.g., Dialogflow requiring sending user data to a third-party server), MindMeld provides robust and fully open-source toolkits that can be deployed as a standalone service. Although, MindMeld provides various functionality for conversational flow this study utilizes only it's dialogue manager functionality. The remaining components like natural language processing and question-answering capabilities are replaced with custom-built modules. The following sections will describe details of data set and method development.

The CA system architecture consists of 3 following components:
\begin{itemize}
    \item The first component as shown in \hyperref[figure1]{Figure 1} below consists of a question understanding module which has 2 sub-modules - question type classifier and named entity recognizer (NER). 
    
    \item Second component is the DS knowledge base - iDISK. We utilize this to provide fast reliable answers to user queries.
    
    \item The third component is answer generation. We use MindMeld \cite{mm} as our base CA framework and within it we have one sub-module
    for answer retrieval, that queries the "Knowledge Base" (KB) - iDISK. This KB is part of our CA framework and contains data extracted from iDISK obtained after performing various transformation and pre-processing steps (discussed in \hyperref[section2.4]{Section 2.4}). Thus, it is analogous to a secondary database or cached data and helps for faster information retrieval.
    \newline
    \newline
   
\end{itemize}

\subsection{Dataset and annotation}
\label{section2.2}

Dataset was accrued by extracting questions (including their titles and contents) from one of the sub-categories “Alternative Medicine” in Yahoo! Answers \cite{yahoo}. We randomly extracted 2000 questions from Yahoo! Answers corpora and manually annotated these samples for pre-defined 12 questions types and 4 named entities (details below). These annotated files were in XMI format which were then cleaned and pre-processed to extract around labelled question samples containing questions type and question focus information (named entities). The standard pre-processing steps included removing duplicates and removing classes which had too few samples (<5) relative to overall class distribution.

Given the variability in quality of the corpus, the question types were narrowed to 8 which were likely to be answered by the iDISK knowledge base and subsequently developed the annotation guidelines. The question types include interaction, usage, effectiveness, adverse effects, indication, background, safety and availability. Initially, three annotators manually annotated a random set of 100 questions for question focus and question type information. The team compared the results and discussed the discrepancies until a consensus was reached. The second iteration included annotation of another set of 100 questions and the inter-annotator agreement was assessed using Cohen’s kappa \cite{cohen_kappa} resulting in a score of 0.86. Henceforth, a final corpus of 2000 questions was annotated for question focus, question type and named entities. However while annotating, questions which were irrelevant and not complete were excluded. Finally, the annotation includes 1509 labelled questions.

\subsection{Question Understanding}
\label{section2.3}

Question understanding is the most essential component for developing our system. Towards question understanding with respect to the domain of DS, there are 2 main tasks  - understanding what type of questions are being asked and recognizing all relevant named entities that the user asks about. The first task can be formulated as question type classification whereas the second task can be viewed as a NER problem.

   \subsubsection{Question Classification} 
    For this classification task, classes are the 8 questions types (discussed in \hyperref[section2.2]{Section 2.2}).
    In this study, we bench-marked 3 machine learning models and 4 deep learning models for question classification using our annotated data (discussed in \hyperref[section2.2]{Section 2.2}). The machine learning models include with Random Forest as a baseline, support vector machine (SVM), and logistic regression. Sparse vector representations for every word are obtained using GloVe embeddings \cite{glove} (840B tokens, 2.2M vocab, cased, 300 dimensions). A comprehensive evaluation of the explored methods is described in \hyperref[section2.5]{Section 2.5}. 
    
    Among deep learning methods, we experimented with Long-Short Term Memory (LSTM), Bi-LSTM, Bi-GRU with attention, Recurrent Neural network (RNN) and Convolutional Neural Network (CNN) models \cite{kim2014convolutional}. The CNN model consists of 4 layers - an embedding layer, a convolutional layer, pooling layers and a full-connected layer with softmax. Vector representations for every word in the sentence is obtained using Glove embeddings. Varied filter sizes of (1,2,3,5) are applied to convolutional layer followed by the max pooling layer which is further used to extract the most important or relevant features generated from the convolution layer. The max pooling scores from each filter were concatenated to form a single vector, which goes through a dropout and is fed into a fully connected layer. LSTM, Bi-LSTM and Bi-GRU are all shallow networks with only 4 layers. The last layer for all the 3 methods is a fully-connected layer with softmax activation.
 
   We experimented with a set of parameters, including the number of filters (ie, 64, 128, 256, and 512), and filter sizes (ie, 1,2,3,5, and 7). The optimal hyperparameters are as follows: positional embedding dimension of 300, filter sizes of 1-7 and, 128 filters for each size. The dropout rate is 0.1,0.2. Weighted precision, recall, and F1 score were used as the evaluation metrics. Due to the limited sample size, we used 10-fold cross validation to evaluate the question classification component.

  \subsubsection{Named Entity Recognition}
    After identifying the question type for a user query, it is essential to recognise what relevant named entities are present in the query. These named entities along with question type guide the answer generation process. The annotated dataset includes 4 types of entities - Dietary Supplement (DS), Disease (DIS), Medication (MED) and Miscellaneous (MISC). These samples were parsed using Spacy dependency parsing \cite{spacy} and following BIO \cite{bio} tags schema.
    
    For the NER task, we experimented with BERT \cite{devlin2019bert} and BiLSTM-CRF \cite{huang2015bidirectional} including a linear statistical model - Hidden Markov Model (HMM) \cite{hmm}. BiLSTM-CRF based methods are standard for NER task. This network can efficiently use past and future input features via a Bi-LSTM layer and sentence level tag information via a CRF layer. A CRF layer is represented by lines which connect consecutive output layers. But recently BERT was among state-of-art methods for the NER task. But for this study, considering the limited dataset size, a simpler model would intuitively be a better choice. This reasoning was also aided by two other factors - faster inference time and class imbalance. Thus, CRF model outperforms other methods in both these aspects. 
    
    The BiLSTM model comprised of 4 layers including a BiLSTM layer and a last CRF layer. Different numbers of hidden units in the Bi-LSTM layer were tested (ie, 30, 64, 128) and the optimal hidden size was set as 30. The Relu activation, dropout of 0.1 and Adam optimiser were used for training the model. Optimal hyper-parameters for best performing vanilla CRF model - c1, c2 as 0.1, 32 batch size, maximum iteration - 100. Features used to train the CRF model include word suffix, part-of-speech (POS) tags, the POS tags of the nearby words (1 word before and 1 word after), etc. Due to the limited sample size, we used 10-fold cross validation to evaluate the question classification component. Weighted precision, recall, F1 and accuracy metrics were used for evaluation.

\subsection{iDISK knowledge base}\label{section2.4}

User questions around DS are answered by retrieving information from the iDISK \cite{idisk}, which integrates and standardizes DS-related information from 4 existing resources including the NM, the “About Herbs” page on the MSKCC website, the DSLD, and the LNHPD. It consists of 7 concept types (i.e., SDSI: Semantic Dietary Supplement Ingredient, DSP: Dietary Supplement Product, DIS: Disease or Syndrome, SPD: Pharmaceutical Drug, SOC: System Organ Class, SS: Sign or Symptom, TC: Therapeutic Class) and 6 relations (i.e., has\_adverse\_effect\_on (SDSI,SOC)’, ‘has\_adverse\_reaction (SDSI, SS)’, ‘has\_ingredient (DSP, SDSI)’, ‘has\_therapeutic\_class (SDSI, TC), ‘interacts\_with (SDSI, SPD)’, ‘is\_effective\_for (SDSI, DIS)’), where the relations entails different relationships between these concept types. The current version of iDISK contains 4,208 DS concepts, which are linked via 6 relationship types to 495 drugs, 776 diseases, 985 symptoms, 605 therapeutic classes, 17 system organ classes, and 137,568 DS products. It is publicly available through \href{https://conservancy.umn.edu/handle/11299/204783}{this link}. Mode details regarding this are discussed in \hyperref[section2.5]{section 2.5} and below.
\newline
\newline
To answer DS related questions, we exploit iDISK knowledge base (details in \hyperref[section2.4]{section 2.4}). This ensures a faster and more accurate domain-oriented query resolution as compared to open-domain answering using the web or search engines. There are various transformations that are performed on our parent database iDISK and these transformations are stored as JSON objects to create MindMeld databases. Search API queries are built to search the knowledge base using Elasticsearch \cite{elasticsearch}. Every domain and subsequent intent can use a separate knowledge base. To query a knowledge base, we have to create a corresponding index which represents that database which in-turn can be called by a particular function. Our framework exploits this functionality as we have more than 3 knowledge base files (JSON) each containing information for every relationship (e.g., has\_ingredient) or domain that is answerable by iDISK. To understand the transformations it would be beneficial to superficially describe the types of questions that can be answered by our CA. Questions pertaining to a) any sort of relations (discussed in \hyperref[section2.4]{section 2.4}), b) background or generic information, c) source information of any DS can be answered. As the questions are diverse within the domain of DS, to answer questions which involve any sort of relationships between two named entities, we perform joins and restrict the number of entries. The join is usually between MRREL.RRF \cite{idisk} which stores the relation (eg, "interacts\_with") and their source information, and MRCONSO.RRF which stores various DS names. Finally, all the data is transformed into JSON format which is fed into our CA framework. As our system also supports one-to-many mapped questions, each entry in the knowledge base can have multiple occurrences in different json dumps. Once the appropriate information is retrieved, we route these to appropriate response/answer handlers. These handlers help generate full semantic responses which are in some form of pre-defined templates.

\subsection{Answer Generation}\label{section2.5}
Results of both question type classification and NER enable us to generate appropriate queries to the knowledge base. The type of question being asked is used to know which relation types to look for and named entities extracted from user query are used to structure appropriate knowledge-base searches. 
To generate natural language responses to user requests, once we get the relevant results from the knowledge base, slot-filling style templates are used which is rendered by a "Responder" object \cite{mm}.


\subsection{Evaluation of the CA system}
This section details the investigative steps performed for each of the CA system components and their corresponding results. We randomly collected another set of 64 examples in addition to the 1509 samples (refer \hyperref[section2.2]{Section 2.2}) and were designated to be used as a hold-out test set for end-user evaluation (This set is not a part of the validation samples in previous section). For evaluating our CA’s responses for a user question. Following the LiveQA Track \cite{livetrack}’s (also followed by Dina et. al \cite{dina}, which simplified the range to a more comprehensive scale) judgement scores on a scale of 1 to 4: 1 - incorrect, 2 - incorrect but related, 3 - correct but incomplete, 4 - correct and complete answer. The final results were then transferred to 0-3 range where 0 for poor or unreadable, 1 for fair, 2 for good and 3 for excellent. With this scale we computed two metrics:
\begin{itemize}
    \item \textit{Average score:} This evaluates the first retrieved answer for every test question (transfers 1–4 level grades to 0–3 scores) \cite{dina}\cite{livetrack}.
    
    \item \textit{succ@i+:} number of questions with score i or above (where i ranges from 2-4) divided by the total number of questions. For example, succ@2+ measures the percent of questions with at least a fair grade answered by the CA \cite{livetrack}.
\end{itemize}

While average score and mean reciprocal rank \cite{mrr} measure how reliable or fair the answers are (evaluated by humans), we also compute Response Error Rate (RER) \cite{ca_comp} as a measure of how coherent and accurate our CA is.
\newline

\section{Results}

\subsection{Question type annotation dataset} 
\hyperref[table1]{Table 1} displays the distribution of 8 question types in our annotated dataset. Question type "Effectiveness" has the highest number of samples whereas "Safety" has the least. The 8 classes are imbalanced and hence we use weighted F1 as discussed \hyperref[section2.5]{section 2.5}. \hyperref[table2]{Table 2} lists sample questions with their annotated question type and named entities.
\begin{table}[ht!]
\label{table1}
\caption{Distribution of samples for 8 classes in our annotated dataset}
  \begin{tabular}{|c|c|}
    \hline
    \textbf{Question Type}  & \textbf{Number of Samples}\\ 
    \hline
     Availability & 147\\
    \hline
     Adverse Effects & 133\\
    \hline
     Background & 125\\
    \hline
     Effectiveness & 318\\
    \hline
     Indication & 188\\
    \hline
     Interaction & 237\\
    \hline
     Safety & 108\\
    \hline
     Usage & 253\\
    \hline
  \end{tabular}
\end{table}

\begin{table}[ht!]
\label{table2}
\caption{Examples of questions in the dataset. Each question is annotated for it's question type and named entity within the question. The column "Named Entity" contains one or two entries. The first entry are entities of type "DS" except the rows marked with $^\star$ and $^\dagger$. The rows with $^\star$ have a second entity entry of type "DIS" and rows with $^\dagger$ have second entry as "MED".}
  \begin{tabular}{|c|c|c|}
    \hline
    \textbf{Question}  & \textbf{Type}   & \textbf{Named Entity} \\ 
    \hline
    Does anyone know if you can take ephedrine while taking levothyroxine? & Interaction & ephedrine, levothyroxine$^\dagger$\\
    \hline
    What is the appropriate dosage of L-glutamine for IBS condition?  & Usage & L-glutamine, IBS$^\star$\\
    \hline
    Does Niacin really work?            & Effectiveness & Niacin\\ 
    \hline
    Does acai berry cause headache?     & Adverse Effects & acai berry, headache$^\star$\\
    \hline
    What are the benefits of using valerian root?   & Indication  & valerian root\\ 
    \hline
    What exactly is ginseng?            & Background & gingseng\\ 
    \hline
    Is kratom safe during pregnancy?    & Safety & kratom\\ 
    \hline
    Where can I buy Selenium pills?     & Availabilty & Selenium\\ 
    \hline
  \end{tabular}
\end{table}

\subsection{Evaluation of Question Understanding Module}
This section details the results of evaluation on the 2 sub-modules of question understanding - question type classification and NER. \hyperref[table3]{Table 3} displays evaluation results of different question classifiers in our question understanding module. The results are reported for 4 metrics - Precision, recall, F1 (all of which are weighted) and accuracy. Our deep learning methods consistently outperform machine learning methods including our baseline model. CNN has the best performance on question classification with weighted F1 score of 81\%. Hence, overall F1 weighted for question classification which is also our standard accuracy is 81\%. \hyperref[table4]{Table 4} includes results for NER submodule. Vanilla CRF model outperforms other 3 methods and has F1 weighted of 85\%.

\begin{table}[ht!]
\caption{Performance of various question type classifiers.}
  \begin{tabular}{|c|c|c|c|c|}
    \hline
    \textbf{Model}  & \textbf{Precision }  & \textbf{Recall} & \textbf{F1} & \textbf{Accuracy}\\ 
    \hline
    Random Forest       & 0.603  & 0.556  & 0.540 & 55.60\%\\ 
    
    \hline
    SVM (with RBF Kernel)    & 0.629  & 0.629 & 0.620 & 62.90\%\\
    \hline
    
    SVM (with Linear Kernel)  & 0.647  & 0.649  & 0.640 & 64.90\%\\
    \hline
    
    Logistic Regression & 0.656  & 0.649  & 0.64 & 64.90\%\\
    \hline

    BiGRU +Attention    & 0.737 & 0.728  & 0.729 & 72.85\%\\ 
    \hline
    
    LSTM                & 0.800   & 0.790   & 0.790 & 78.67\%\\ 
    \hline
    Bi-LSTM             & 0.800  & 0.790    & 0.790 & 79.41\%\\ 
    \hline

    \textbf{CNN}        & \textbf{0.810}  & \textbf{0.800}    & \textbf{0.810} & \textbf{80.79}\%\\ 
    \hline
  \end{tabular}
  \label{table3}
\end{table}

\begin{table}[ht!]
\caption{Performance of various entity recognizers.}
  \begin{tabular}{|c|c|c|c|c|}
    \hline
    \textbf{Model}  & \textbf{Precision }  & \textbf{Recall} & \textbf{F1} & \textbf{Accuracy}\\ 
    \hline
    BERT    & 0.629  & 0.629 & 0.620 & 62.90\%\\
    \hline
    BiLSTM+CRF & 0.723  & 0.678  & 0.698 & 73.20\%\\
    \hline
    HMM      & 0.719  & 0.820  & 0.740 & 82.00\%\\
    \hline
    \textbf{CRF}     & \textbf{0.872}   & \textbf{0.860}   & \textbf{0.850}  & \textbf{87.10}\%\\
    \hline
  \end{tabular}
  \label{table4}
\end{table}

\subsection{Evaluation of responses for user questions}
\hyperref[Table5] {Table 5} displays a sample of responses from our CA. On our evaluation set of 64 question-answer pairs, we observed an average score of 1.82 that indicates that responses are fair and mostly good. For the succ@i metric we compute succ@2+ and succ@3+ and observe scores of 0.762 and 0.656 respectively. These scores are consistent with the intuition that as we increment the score range (grade scale 1-4) probability of correct answers decreases. Thus, the system is able to answer more than 76.2\% of the questions in a fair, relatable way and more than 66\% in a good way. RER metric returns a score of 0.23 which indicates that the answers are reasonably coherent.

\begin{table}[!ht]
\caption{Examples of Questions and CA system responses}
\setlength\extrarowheight{2pt} 
\begin{tabularx}{\textwidth}{|C|C|}
\hline
\textbf{Question} & \textbf{Answer}  \\\hline
is it safe to take melatonin? & Here is what I found about pure crystalline Melatonin: Melatonin may cause drowsiness. Patients should not drive or operate heavy machinery until familiar with the effects of melatonin.\\
\hline
are there any dangerous side effects that anyone has experienced with the supplement milk thistle? & The Blessed Thistle preparation has adverse effects like Eye disorders.\\
\hline
are there any proven benefits to taking shark cartilage? & Shark Cartilage is effective for Degenerative Polyarthritis.\\
\hline
\end{tabularx}
\label{Table5}
\end{table}


\section{Discussion}
This paper describes the first CA system for answering DS questions to consumers. Developing specialized domain restricted CA systems is a challenging task and especially for a largely unexplored domain such as DS. In this paper, we presented our work over building a robust CA for answering real-world questions on DS collected from Yahoo! Answers. 

Question understanding can be formulated into 2 subtasks, each of which is discussed in \hyperref[section2.3]{Section 2.3}. For question type classification task, despite restricted dataset, deep learning methods outperformed conventional machine learning models. Specifically, among different deep learning methods, CNN based question type classification gave best results. 
This is primarily because a CNN architecture is able to incorporate some local context. CNN outperforms shallow models such as SVM, Naive Bayes because CNN can provide semantically meaningful feature representations as compared to shallow learning models \cite{li2020survey}. But, CNN is also more accurate than a LSTM, especially during the feature extraction step because a 1D CNN \cite{kim2014convolutional} processes text as a one-dimensional image and is used to capture the latent associations between neighboring words (spatial context), in contrast with LSTMs, which process each word in a sequential pattern \cite{jang_new}\cite{liu}. Thus, many works like \cite{jang_new} exploit an ensemble model of CNN and LSTM.  

For our second task - NER, it is unsurprising that a single CRF layer achieves the best results given our limited data points. (\hyperref[section2.3]{Section 2.3}). Comparative performance of CRF and LSTM does not follow a particular trend but dependent on several crucial factors, mainly size and type of training data which is similar to factors we discussed just now. In \cite{berg-dalianis-2019-augmenting}, LSTM with CRF performs reasonably well by itself, although, the CRF model seems to perform better. As shown in studies \cite{leevy}\cite{dernoncourt2016deidentification}, LSTM approach performs best. Also, there is no conclusive evidence that a combined use of LSTM and CRF systems in hybrid or ensemble models will always outperform vanilla CRF or LSTM, such as \cite{dernoncourt2016deidentification}. Thus, suitability of a model architecture in most cases is use-case dependent and for our case, due to additional factors such as time constraints (\hyperref[section2.3]{Section 2.3}), CRF approach works the best.

Both extracted question type and named entities help the CA to generate appropriate responses to a user query. Question classification output tells us which type of question we are dealing with, whereas NER module extracts relevant named entities present in the query which helps the CA to give most suitable result from the knowledge base.  We described our end-to-end evaluation method in section 2.6 which aims to ensure the feasibility of the whole system. Query response was evaluated using average score and succ@i and robustness of our conversational system as a whole was evaluated using RER. The system achieves an overall accuracy of 81\% and an average score of 1.82 with succ@3+ score as 76.2\% and succ@2+ as 66\% approximately. The succ@3+ score is greater than succ@2+ score, which is an expected because as we  increment  the  score  range  (grade  scale1-4) probability of correct answers decreases.

Error analysis with respect to the RER score showed a possible explanation for a relatively higher score is discrepancy individual in individual reviewers' preference. Some preferred long detailed answers where as some gave more preference to crisp responses. This was reflected in the score ranges of many answers. Also, among our test samples for few questions we saw wrong answers being generated by our system and were rated 1 by the reviews. These wrong answers were mainly a result of error propagation from either the NER step or the question classification step. For example, our system was unable to identify cinnamon sticks as an entity in the question "does eating cinnamon sticks really get rid of a uti?" and consequently wrongly answered the question. Thus, both these factors contributed towards a higher RER on average but in totality the system still remained reasonably coherent in its performance.   

Our work has several limitations. The sample size is still very limited. The domain coverage is not quite wide and focuses only on three major relations from iDISK. We still need to find a more intensive way to categorize the questions types into not the 8 predefined classes but a mix of them as well so that layered questions which span more than single question types can be answered more reliably. We will also expand hierarchical components of intents and entities further along with respective annotated samples so that we can increase generalization of questions within confines of our DS domain. More comprehensive experimentation (with more data) is required to evaluate the generated responses and their refinement into more human-like answers. We should also incorporate user feedback when the system is fully developed. The future scope of this work is to explore development of word embeddings specific to DS domain and introduce voice-based input for the conversational system. Unlike many other systems using retrieval-based methods, we leverage backend knowledge base, as it provides high-quality and easy-to-query knowledge. In addition, DS information is disparate through the Internet, and availability and quality of question-answer pairs are uncertain.

\section{Conclusion}
In this study, we have developed a CA system for answering user questions about DS use using a DS knowledge base and MindMeld framework. We developed two components for understanding the natural language questions: question type classifier (F1: 0.810) and named entity recognition (F1 0.850). We demonstrated that it is feasible to integrated developed models into MindMeld framework and answer most questions accurately. 

\section*{Appendix}

\begin{backmatter}

\section*{Acknowledgements}
We would like to acknowledge and thank Andrew Yang and Andrew Wang for their contribution on annotations and evaluation of the system.

\section*{Funding}
National Institutions of Health’s National Center for Complementary Integrative Health (NCCIH), the Office of Dietary Supplements (ODS) and National Institute on Aging (NIA) grant number R01AT009457 (PI: Zhang) 

\section*{Abbreviations}
\begin{itemize}
    \item AI - Artificial Intelligence
    \item API - Application Programming Interface
    \item Bi-LSTM - Bi-directional LSTM
    \item BIO - Beginning, Inside, Out
    \item CA - Conversation System
    \item CRF - Conditional Random Fields
    \item CNN - Convolutional Neural Network
    \item DIS - Disease
    \item DL - Deep Learning
    \item DS - Dietary Supplement
    \item KB - Knowledge-base
    \item MED - Medication
    \item MISC - Miscellaneous 
    \item ML - Machine Learning
    \item NER - Named Entity Recognition
    \item NLP - Natural Language Processing
    \item NM - Natural Medicine
    \item RER - Response Error Rate
    \item QU - Question Understanding
    \item XMI - XML Metadata Interchange

\end{itemize}



\section*{Competing interests}
The authors declare that they have no competing interests.





\bibliographystyle{bmc-mathphys} 
\bibliography{bmc_article}      

\end{backmatter}
\end{document}